\documentclass[10pt,twocolumn,letterpaper]{article}
\usepackage{iccv}
\usepackage{times}
\usepackage{epsfig}
\usepackage{graphicx}
\usepackage{amsmath}
\usepackage{amssymb}
\usepackage{commath}
\usepackage{boldline}
\usepackage{float}
\usepackage{flushend}
\usepackage[pagebackref=true,breaklinks=true,letterpaper=true,colorlinks,bookmarks=false]{hyperref}

\iccvfinalcopy % *** Uncomment this line for the final submission

\def\iccvPaperID{****} % *** Enter the ICCV Paper ID here

\newcommand{\eat}[1]{}

\begin{document}
%%%%%%%%% TITLE
\title{3D Object Reconstruction from a Single Depth View
with Adversarial Learning 
}

\author{Bo Yang\\
University of Oxford\\
{\tt\small bo.yang@cs.ox.ac.uk}
\and
Hongkai Wen\\
University of Warwick\\
{\tt\small hongkai.wen@dcs.warwick.ac.uk }
\and
Sen Wang\\
Heriot-Watt University\\
{\tt\small s.wang@hw.ac.uk}
\and
Ronald Clark\\
Imperial College London\\
{\tt\small ronald.clark@imperial.ac.uk}
\and
Andrew Markham\\
University of Oxford\\
{\tt\small andrew.markham@cs.ox.ac.uk}
\and
Niki Trigoni\\
University of Oxford\\
{\tt\small niki.trigoni@cs.ox.ac.uk}
}

\makeatletter
\g@addto@macro\@maketitle{
  \vspace{-1 cm}
  \begin{figure}[H]
  \setlength{\linewidth}{\textwidth}
  \setlength{\hsize}{\textwidth}
   \setlength{\belowcaptionskip}{ -12pt}
  \centering
  \includegraphics[width=0.8\textwidth]{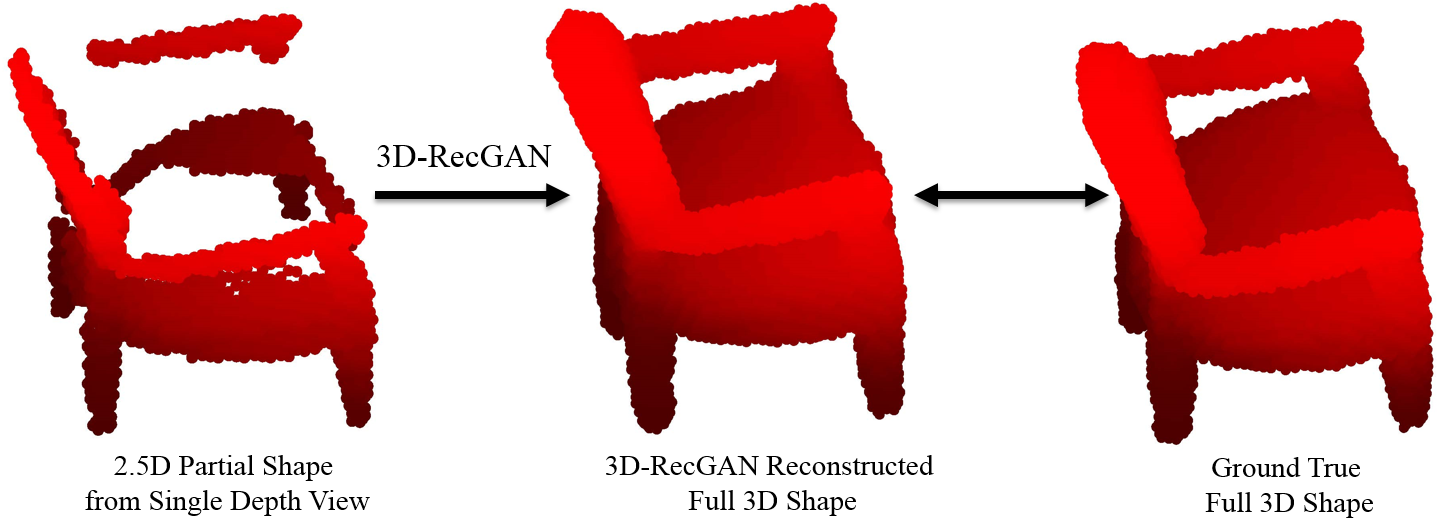}
  \caption{Our method 3D-RecGAN reconstructs a full 3D shape from a single 2.5D depth view.}
  \label{fig:rec_sample}
  \end{figure}
}
\makeatother
\maketitle
%\thispagestyle{empty}

%%%%%%%%% ABSTRACT
\begin{abstract}
\vspace{-0.4 cm}
In this paper, we propose a novel \textbf{3D-RecGAN} approach, which reconstructs the complete 3D structure of a given object from a single arbitrary depth view using generative adversarial networks. Unlike the existing work which typically requires multiple views of the same object or class labels to recover the full 3D geometry, the proposed 3D-RecGAN only takes the voxel grid representation of a depth view of the object as input, and is able to generate the complete 3D occupancy grid by filling in the occluded/missing regions. The key idea is to combine the generative capabilities of autoencoders and the conditional Generative Adversarial Networks (GAN) framework, to infer accurate and fine-grained 3D structures of objects in high-dimensional voxel space. Extensive experiments on large synthetic datasets show that the proposed 3D-RecGAN significantly outperforms the state of the art in single view 3D object reconstruction, and is able to reconstruct unseen types of objects. Our code and data are available at:  \url{https://github.com/Yang7879/3D-RecGAN}.
\end{abstract}

\vspace{-0.4 cm}
\section{Introduction}\label{sec:intro}
The ability to reconstruct the complete and accurate 3D geometry of an object is essential for a %\hongkai{
broad spectrum of scenarios, from AR/VR applications \cite{B2016e} and semantic understanding, to robot grasping \cite{Varley2017} and obstacle avoidance. One class of popular approaches is to use the off-the-shelf low-cost depth sensing devices such as Kinect and RealSense cameras to recover the 3D model of an object from captured depth images. Most of those approaches typically sample multiple depth images from different views of the object to create the complete 3D structure \cite{Newcombe2011a} \cite{Niener2013} \cite{Steinbr2013}. However, in practice it is not always feasible to scan all surfaces of the object, which leads to incomplete models with occluded regions and large holes. In addition, acquiring and processing multiple depth views require significant computational power, which is not ideal in many applications that require real-time response.

In this paper, we aim to tackle the problem of inferring the complete 3D model of an object using a single depth view. This is a very challenging task, since the partial observation of the object (i.e. a depth image from one viewing angle) can be theoretically associated with infinite number of possible 3D models. Traditional reconstruction approaches typically use interpolation techniques such as plane fitting \cite{Sorkine2004} or Poisson surface estimation \cite{Kazhdan2006} \cite{Kazhdan2013} to estimate the underlying 3D structure. However, they can only recover very limited occluded/missing regions, e.g. small holes or gaps due to quantization artifacts, sensor noise and insufficient geometry information. 

Interestingly, humans are surprisingly talent at such ambiguity by implicitly leveraging prior knowledge. For example, given a view of a  chair with two rear legs occluded by front legs, humans are easily able to guess the most likely shape behind the visible parts. Recent advances in deep neural nets and data driven approaches are suitable to deal with such a task. 

In this paper, we aim to acquire the complete 3D geometry of an object given a single depth view. By utilizing the high performance of 3D convolutional neural nets and large open datasets of 3D models, our approach learns a smooth function to map a 2.5D view to a complete 3D shape. Particularly, we train an end-to-end model which estimates full volumetric occupancy from only one 2.5D depth view of an object, thus predicting occluded structures from a partial scan.  

While state-of-the-art deep learning approaches \cite{Dai2017b} \cite{Wu2015} \cite{Chan2016} \cite{Varley2017} \cite{Yan2016} for 3D shape reconstruction achieve encouraging and compelling results, they are limited to a very small resolution, typically less than $40^3$ voxel grids. As a result, the learnt 3D shape tends to be coarse and inaccurate. However, to increase the model resolution without sacrificing recovery accuracy is challenging, as even a slightly higher resolution would exponentially increase the search space of potential 2.5D to 3D mapping functions, resulting in difficulties in convergence of neural nets.

Recently, deep generative models achieve impressive success in modeling complex high-dimensional data distribution, among which Generative Adversarial Networks (GANs) \cite{Goodfellow2014} and Variational Autoencoders (VAEs) \cite{Kingma2014} emerge as two powerful frameworks for generative learning, including image and text generation \cite{Radford2016} \cite{Hu2017a}, and latent space learning \cite{Chen2016a} \cite{Kulkarni2015}. In the past two years, a number of works \cite{Girdhar} \cite{Wu2016a} \cite{B2016f} \cite{Huang2015a} apply such generative models to learn latent space to represent 3D object shapes, and then to solve simple discriminative tasks such as new image generation, object classification, recognition and shape retrieval. However, 3D shape reconstruction, as a more difficult generative task, has yet to be fully explored. 

In this paper, we propose 3D-RecGAN, a novel model that combines both an autoencoder and GAN to generate a full 3D structure conditioned on a single 2.5D view. Particularly, our model first encodes the 2.5D view to a low-dimensional latent space vector which implicitly represents general 3D  geometric structures, then decodes it back to recover the most likely complete 3D structure. The rough 3D structure is then feed into a conditional discriminator which is adversarially trained to distinguish whether the coarse 3D shape is plausible or not.The autoencoder is able to approximate the corresponding shape, while the adversarial training tends to add fine details to the estimated shape. To ensure the final generated 3D shape corresponds to the input single partial 2.5D view, adversarial training of our model is based on conditional GAN \cite{Mirza2014} instead of random guessing. 

Our contributions are as follows:

(1) We formulate a novel generative model to reconstruct the full 3D structure using a single arbitrary depth view. By drawing on both autoencoder and GAN, our approach is end-to-end trainable with high level of generality. Particularly, our model consumes a simple occupancy grid map without requiring object class labels or any annotations, while predicting a compelling shape with a high resolution of $64^3$ voxel grid.

(2) We exploit conditional GAN during training to refine 3D shape estimates from autoencoder. Key contribution here is the use of a latent distribution rather than a binary variable from the discriminator to train both discriminator and autoencoder. Using a latent distribution of high-dimensional real or fake 3D reconstructed shapes from discriminator significantly stabilizes the training of GAN, while using the standard binary variable 0/1 for training leads to the GAN crash easily.

(3) We conduct extensive experiments for single category and multi-category reconstruction, outperforming the state of the art. Besides, our approach is also able to generalize previously unseen object categories.

%Our main contribution is to formulate an end-to-end trainable 3D convnet model for 3D reconstruction from single arbitrary view. To the best of our knowledge, this work is among the first to directly leverage conditional GAN for the challenging 3D shape completion. Compared with the state-of-the-art, our 3D-RecGAN predicts higher resolution of $64^3$ occupancy grid, while others \cite{Dai2017b} \cite{Chan2016} \cite{Varley2017} \cite{Wu2015} \cite{Yan2016} \cite{Rezende2016} \cite{Gadelha2016} \cite{B2016e} only generate less than $40^3$ grids. Besides, the proposed model only consumes a simple occupancy grip map, while many other works rely on more complicated map encodings such as truncated signed distance field (TSDF) \cite{Curless1996}\cite{Dai2017b}, flipped TSDF \cite{Song2017}  and Silhouettes\cite{B2016h}. Unlike previous works, our model does not require object class labels or any annotations. At last, we systematically evaluate 3D shape recovery in three levels: per-category, multi-category and cross-category. 

We evaluate our approach on synthetic datasets from virtually scanned 3D CAD models. Ideally, this task should be evaluated on real world 2.5D depth views, but it is very challenging to obtain the ground truth of 3D shape with regard to a specific 2.5D view for both training and evaluation. To the best of our knowledge, there are no good open datasets which have the ground truth for occluded/missing parts and holes for each 2.5D view in real world. Extensive experiments demonstrate that our 3D-RecGAN outperforms the state of the art by a large margin. Our reconstruction results are not only quantitatively more accurate, but also qualitatively with more details. An example of chair completion is shown in Figure \ref{fig:rec_sample}.

\section{Related Work}\label{sec:liter}
We review different pipelines for 3D reconstruction or shape completion. Both conventional geometry based and the state-of-the-art deep learning based approaches are covered.

(1) \textbf{3D Model/Shape Fitting}. \cite{Monszpart2015a} uses plane fitting to complete small missing regions, while \cite{Mattausch2014} \cite{Mitra2006} \cite{Pauly2008} \cite{Sipiran2014} \cite{Speciale2016} \cite{Thrun2005} applies shape symmetry to fill in holes. Although these methods show good results, relying on predefined geometric regularities fundamentally limits the structure space to hand-crafted shapes. Besides, these approaches are likely to fail when missing or occluded regions are relatively big. Another similar fitting pipeline is to leverage database priors. Given a partial shape input, \cite{Kim2012} \cite{Li2015a} \cite{Nan} \cite{Shao2012} \cite{Shi2016} try to retrieve an identical or most likely CAD model and align it with the partial scan. However, these approaches explicitly assume the database contains identical or very similar shapes, thus being unable to generalize novel objects or categories. 

(2) \textbf{Multi-view Reconstruction}. Traditionally, 3D dense recovery requires a collection of images \cite{Hartley2004}. Geometric shape is recovered by dense feature extraction and matching \cite{Newcombe2011b}, or by directly minimizing reprojection errors \cite{Baker2004}. Basically, these methods are used for traditional SfM and visual SLAM, which is unable to build 3D structures for featureless regions such as white walls. Recently, \cite{Gadelha2016} \cite{Rezende2016} \cite{Tulsiani2017} \cite{Tatarchenko2016} \cite{B2016h} \cite{Chan2016} \cite{Riegler2017} \cite{Soltani2017} \cite{Lun2017} leverage deep neural nets to learn a 3D shape from multiple images. Although most of them do not directly require 3D ground-truth labels for supervision during training, they rely on additional signals such as contextual or camera information to supervise the view consistency. Obviously, extra efforts are required to acquire such additional signals. Additionally, resolution of the recovered occupancy shape is usually up to a small scale of $32^3$.

(3) \textbf{Single-view Reconstruction}. Predicting a complete 3D object model from a single view is a long-standing and very challenging task. When reconstructing a specific object category, model templates can be used. For example, morphable 3D models are exploited for face recovery \cite{Blanz2003} \cite{Dou2017}. This concept was extended to reconstruct simple objects in \cite{Kar2015a}. For general and complex object completion, recent machine learning approaches achieve promising results. Firman et al. \cite{Firman2016} trained a random decision forest to predict unknown voxels. 3D ShapeNets \cite{Wu2015} is amongst the early work using deep networks to predict multiple 3D solutions from a single partial view. Fan et al. \cite{Fan2017} also adopted a similar strategy to generate multiple plausible 3D point clouds from a single image. However, that strategy is significantly less efficient than directly training an end-to-end predictor \cite{Dai2017b}. VConv-DAE \cite{B2016e} can be used for shape completion, but it is originally designed for shape denoising rather than partial range scans. Wu et al. proposed 3D-INN \cite{B2016d} to estimate a 3D skeleton from single image, which is far from recovering an accurate and complete 3D structure. Dai et al. developed 3D-EPN \cite{Dai2017b} to complete an object's shape using deep nets to both predict a $32^3$ occupancy grid and then synthesize a higher resolution model based on a shape database. While it achieves promising results, it is not an end-to-end system and it relies on a prior model database. Perspective Transformer Nets \cite{Yan2016} and the recent WS-GAN \cite{Gwak2017} are introduced to learn 3D object structures up to a $32^3$ resolution occupancy grid. Although they do not need explicit 3D labels for supervision, it requires a large number of 2D silhouettes or masks and specific camera parameters. In addition, the training procedure of \cite{Yan2016} is two-stage, rather than end-to-end. Song et al. \cite{Song2017} proposed SSCNet for both 3D scene completion and semantic label prediction. Although it outputs a high resolution occupancy map, it requires strong voxel-level annotations for supervision. It also needs special map encoding techniques such as elimination of both view dependency and strong gradients on TSDF. \cite{Tatarchenko2017} \cite{Riegler2017} use tree structures, while \cite{Guizilini} applies Hibert Maps for 3D map representation to recover the 3D shape, thus being able to produce a relatively higher resolution of 3D shape. However, their deep networks only consist of a 3D encoder and decoder, without taking advantage of adversarial learning. Varley et al. \cite{Varley2017} provides an architecture for 3D shape completion from a single depth view, producing an up to $40^3$ occupancy grid. Although reconstruction results are encouraging, the network is not scalable to higher resolution 3D shape because of the heavy fully connected layers.

\section{3D-RecGAN} \label{sec:net_view}
\subsection{Overview}
\begin{figure}[b]
    \vspace{-0.5 cm}
	\setlength{\abovecaptionskip}{ 0 cm}
    \setlength{\belowcaptionskip}{ -8pt}
    \centering
    \includegraphics[width=0.45\textwidth]{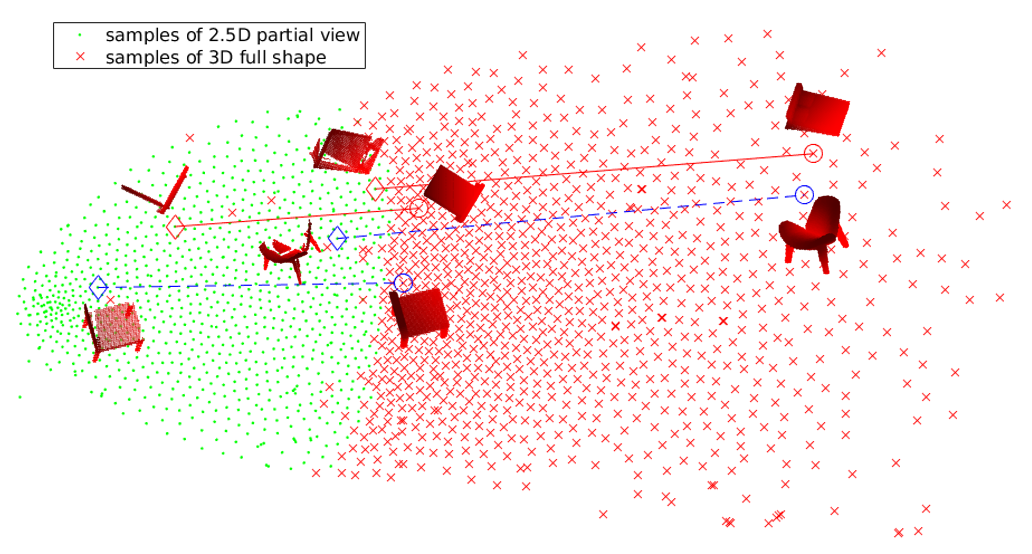}
    \caption{t-SNE embeddings of 2.5D partial views and 3D complete shapes of multiple object categories.}
    \label{fig:t_SNE_multi}
\end{figure}
Our method aims to predict a complete 3D shape of an object, which takes only an arbitrary single 2.5D depth view as input. The output 3D shape is automatically aligned with the corresponding 2.5D partial scan. To achieve this task, each object model is represented in a 3D voxel grid. We only use the simple occupancy information for map encoding, where 1 represents an occupied cell and 0 remains an empty cell. Specifically, both the input, denoted as $I$, and output 3D shape, denoted as $Y$, are $64^3$ occupancy grids in our networks. The input shape is directly calculated from a single depth image. To generate ground true training and evaluation pairs, we virtually scan 3D objects from ModelNet40 \cite{Wu2015}. Figure \ref{fig:t_SNE_multi} is the t-SNE visualization of partial 2.5D views and the corresponding full 3D shapes for multiple general chair and bed models.
Each green dot represents the t-SNE embedding of a 2.5D view, whilst a red dot is the embedding of corresponding 3D shapes. It can be seen that multiple categories inherently have similar 2.5D to 3D mapping relationships. Essentially, our neural network is to learn a smooth function, denoted as $f$, which maps green dots to red dots in high dimensional space as shown in Equation \ref{eq:f25d3d}. The function $f$ is parametrized by convolutional layers in general. 
\begin{figure}[t]
	\setlength{\abovecaptionskip}{ 0 cm}
    \setlength{\belowcaptionskip}{ -8pt}
    \centering
    \includegraphics[width=0.45\textwidth]{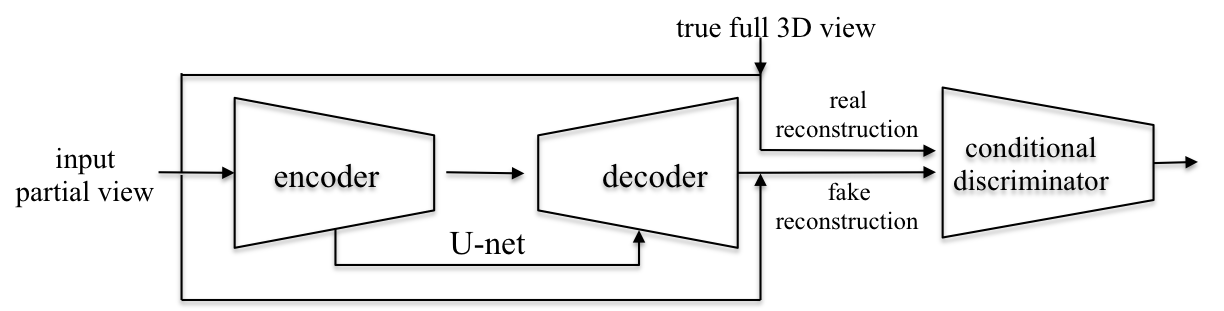}
    \caption{Overview of network architecture for training.}
    \label{fig:net_view_train}
\end{figure}
\begin{figure}[t]
	\setlength{\abovecaptionskip}{ 0 cm}
    \setlength{\belowcaptionskip}{ -12pt}
    \centering
    \includegraphics[width=0.4\textwidth]{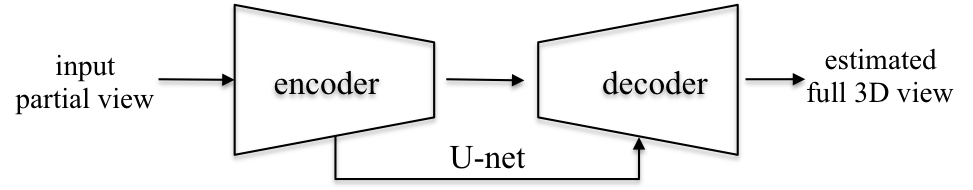}
    \caption{Overview of network architecture for testing.}
    \label{fig:net_view_test}
\end{figure}
\begin{equation}
\label{eq:f25d3d}
Y = f(I) \quad \left( I,Y \in Z_{2}^{64^3}, where {\ } Z_{2}=\{0,1\} \right)
\end{equation}

After generating training pairs, we feed them into our networks. The first part of our network loosely follows the idea of an autoencoder with U-net architecture \cite{Ronneberger2015}. The autoencoder serves as a generator which is followed by a conditional discriminator \cite{Mirza2014} for adversarial learning. %as shown in Figure \ref{fig:net_arch}. 
Instead of reconstructing the original input and learning an efficient encoding, the autoencoder in our network aims to learn a correlation between partial and complete 3D structures. With the supervision of complete 3D labels, the autoencoder is able to learn a function $f$ and generate a reasonable 3D shape given a brand new partial 2.5D view. In the testing phase, however, the results tend to be graining and without fine details. 

To address this issue, in the training phase, the reconstructed 3D shape from the autoencoder is further fed into a conditional discriminator to verify its plausibility. In particular, a partial 2.5D input view is paired with its corresponding complete 3D shape, which is called the ``real reconstruction", while the partial 2.5D view is paired with its corresponding output 3D shape from autoencoder, which is called ``fake reconstruction". The discriminator aims to discriminate all ``fake reconstruction" against ``real reconstruction". In the original GAN framework \cite{Goodfellow2014}, the task of discriminator is to simply classify real and fake input, but its Jensen-Shannon divergence-based loss function is difficult to converge. The recent WGAN \cite{Arjovsky2017} leverages Wasserstein distance with weight clipping as a loss function to stabilize the training procedure, whilst the extended work WGAN-GP \cite{Gulrajani2017} further improves the training process using a gradient penalty with respect to its input. In our 3D-RecGAN, we apply WGAN-GP as the loss function of our conditional discriminator, which guarantees fast and stable convergence. The overall network architecture for training is shown in Figure \ref{fig:net_view_train}, while the testing phase only needs the well trained autoencoder as shown in Figure \ref{fig:net_view_test}.

Overall, the main challenge of 3D reconstruction from an arbitrary single view is to generate new information including filling the missing and occluded regions from unseen views, while keeping the estimated 3D shape corresponding to the specific input 2.5D view. In the training phase, our 3D-RecGAN firstly leverages the autoencoder to generate a reasonable ``fake reconstruction", then applies adversarial learning to refine the ``fake reconstruction" to make it as similar to ``real reconstruction" through jointly updating parameters of autoencoder. In the testing phase, given a novel 2.5D view as input, the jointly trained autoencoder is able to recover a full 3D model with satisfactory accuracy, while the discriminator is no longer used.

\subsection{Architecture } \label{sec:net}
\begin{figure*}[t]
	\setlength{\abovecaptionskip}{ 0 cm}
    \setlength{\belowcaptionskip}{ -8pt}
    \centering
    \includegraphics[width=1\textwidth]{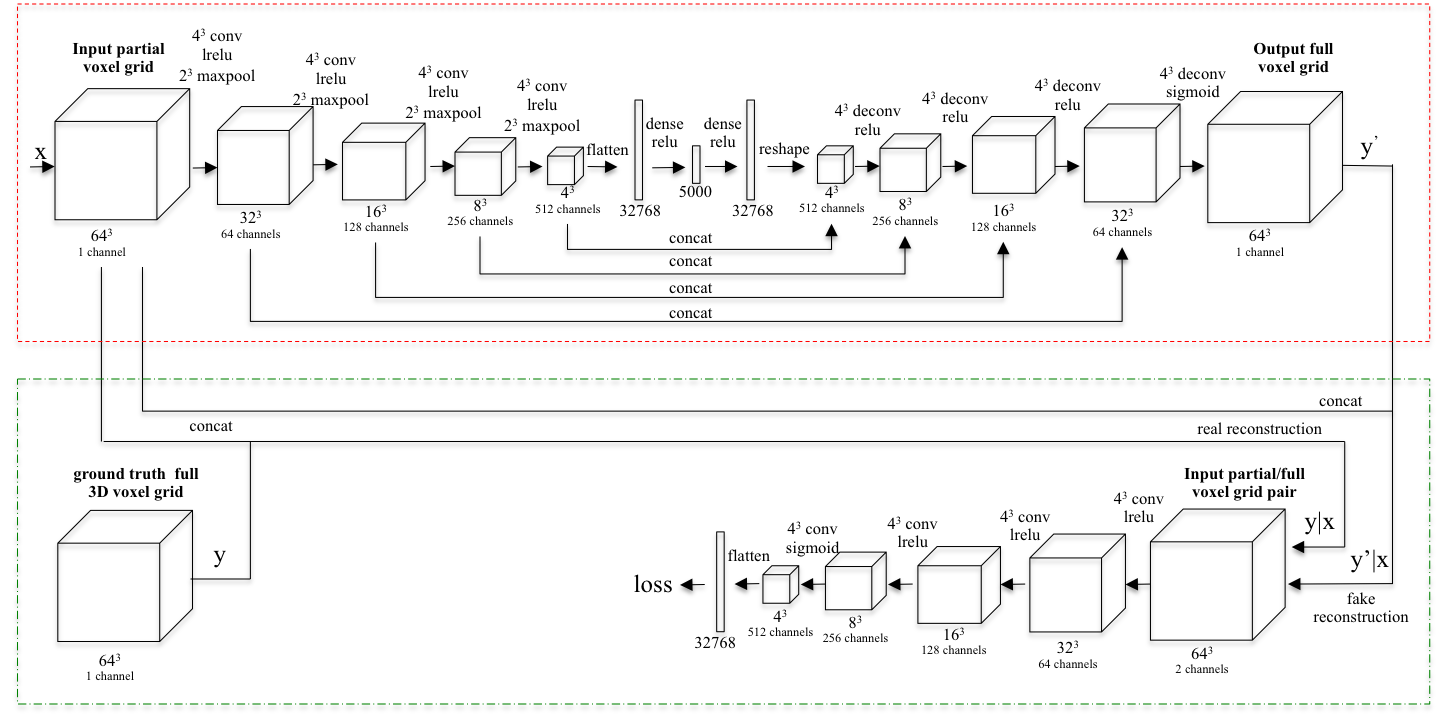}
    \caption{3D-RecGAN Architecture.}
    \label{fig:nets}
\end{figure*}
Figure \ref{fig:nets} shows the detailed architecture of our proposed 3D-RecGAN. It consists of two main networks: the generator as in the top block and the discriminator as in the bottom block. 

\textbf{The generator} is based on autoencoder with skip-connections between encoder and decoder. Unlike the vanilla GAN generator which generates data from arbitrary latent distributions, our 3D-RecGAN generator synthesizes data from latent distribution of 2.5D views. Particularly, the encoder has five 3D convolutional layers, each of which has a bank of 4x4x4 filters with strides of 1x1x1, followed by a leaky ReLU activation function and a max pooling layer which has 2x2x2 filters and strides of 2x2x2. The number of output channels of max pooling layer starts with 64, doubling at each subsequent layer and ends up with 512. The encoder is lastly followed by two fully-connected layers to embed semantic information into latent space. The decoder is composed of 5 symmetric up-convolutional layers which are followed by ReLU activations except for the last layer with sigmoid function. Skip-connections between encoder and decoder guarantee propagation of local structures of the input 2.5D view. It should be noted that without the two fully connected layers and skip-connections, the vanilla autoencoder is unable to learn reasonable full 3D structures as the latent space is limited and the local structure is not preserved. During training, the generator is supervised supplying by ground true 3D shapes. The loss function and optimization methods are described in Section \ref{sec:obj}.

\textbf{The discriminator} aims to distinguish whether the estimated 3D shapes are plausible or not. Based on conditional GAN, the discriminator takes both real reconstruction pairs and fake reconstruction pairs as input. Particularly, it consists of five 3D convolutional layers, each of which has a bank of 4x4x4 filters with strides of 2x2x2, followed by a ReLU activation function except for the last layer which is followed by a sigmoid activation function. The number of output channels of each layer is the same as that in the encoder part. Unlike the original GAN and conditional GAN, our discriminator is not designed as a binary discriminator to simply classify fake against real reconstructions. The reason is both real reconstruction pairs and fake reconstruction pairs are extremely high dimensional distributions, i.e. $2*64^3$ dimensions. To naively classify it as only two categories would result in it being unable to capture geometric details of the object, and the discrimination loss is unlikely to benefit the generator through back-propagation. Instead, our discriminator is designed to output a long latent vector which represents distributions of real and fake reconstructions. Therefore, our discriminator is to distinguish the distributions of latent representations of fake and real reconstructions, while the generator is trained to make the two distributions as similar as possible. We use WGAN-GP as loss functions for our 3D-RecGAN. %Details will be discussed in section \ref{sec:obj}.

\subsection{Objectives } \label{sec:obj}
The objective function of our 3D-RecGAN includes two main parts: an object reconstruction loss $L_{ae}$ for autoencoder based generator; the objective function $L_{gan}$ for conditional GAN.

(1) \textbf{$L_{ae}$} \quad For the generator, inspired by the work \cite{Brock2016}, we use modified binary cross-entropy loss function instead of the standard version. The standard binary cross-entropy weights both false positive and false negative results equally. However, most of the voxel grid tends to be empty and the network easily gets a false positive estimation. In this regard, we impose a high penalty on false positive than false negative results. Particularly, a weight hyperparameter $\alpha$ is assigned to false positives, with $(1-\alpha)$ for false negative results, as shown in following Equation \ref{eq:lae}.
\begin{equation}
\label{eq:lae}
L_{ae} = -\alpha  y \log(y^{'}) - (1-{\alpha} ) (1-y) \log(1-y^{'})
\end{equation}
where $y$ is the target value in \{0,1\} and $y^{'}$ is the estimated value in (0,1) for each voxel from the autoencoder.  

(2) \textbf{$L_{gan}$} \quad For the discriminator, we leverage the state-of-the-art WGAN-GP loss functions. Unlike the original GAN loss function which presents an overall loss for both real and fake inputs, we separately represent the loss function $L_{gan}^{g}$ in Equation \ref{eq:lgang} for generating fake reconstruction pairs and $L_{gan}^{d}$ in Equation \ref{eq:lgand} for discriminating fake and real reconstruction pairs. Detailed definitions and derivation of the loss functions can be found in \cite{Arjovsky2017} \cite{Gulrajani2017}, but we modify them for our conditional GAN settings.  
\begin{equation}
\label{eq:lgang}
L_{gan}^{g} = -\mathbf{E}\left[ D(y^{'}|x) \right]
\end{equation}
\begin{align}
\label{eq:lgand}
L_{gan}^{d} = \mathbf{E}\left[ D(y^{'}|x)\right] - \mathbf{E} \left[ D(y|x) \right] \nonumber \\
+ \lambda \mathbf{E} \left[ \left(\norm{\nabla_{\hat{y}} D(\hat{y}|x)}_2 - 1 \right)^2  \right]
\end{align}
where $\hat{y} = \epsilon x +(1-\epsilon) y^{'}, \epsilon \sim U[0,1]$.  $\lambda$ controls the trade-off between optimizing the gradient penalty and the original objective in WGAN, $x$ represents a voxel value, e.g.\{0,1\}, of an input 2.5D view, while $y^{'}$ is the estimated value in (0,1) for the corresponding voxel from generator, and $y$ is the target value in \{0,1\} for the same voxel.

For the generator in our 3D-RecGAN network, there are two loss functions, $L_{ae}$ and $L_{gan}^{g}$, to optimize. As we discussed in Section \ref{sec:net_view}. Minimizing $L_{ae}$ tends to learn the overall 3D shapes, whilst minimizing $L_{gan}^{g}$ estimates more plausible 3D structures conditioned on input 2.5D views. To minimize $L_{gan}^{d}$ is to improve the performance of discriminator to distinguish fake and real reconstruction pairs.
To jointly optimize the generator, we assign weight $\beta$ to $L_{ae}$, $(1-\beta)$ to $L_{gan}^{g}$. Overall, the loss functions for generator and discriminator are as follows:
\begin{equation}
L_g = \beta L_{ae} + (1-\beta) L_{gan}^{g}
\end{equation}
\begin{equation}
L_d = L_{gan}^{d}
\end{equation}

\subsection{Training } \label{sec:train}
We adopt an end-to-end training procedure for the whole network. To simultaneously optimize both generator and discriminator,  we alternate between one gradient descent step on discriminator and then one step on generator. For the WGAN-GP, $\lambda$ is set as  10 for gradient penalty as in \cite{Gulrajani2017}. $\alpha$ ends up as 0.85 for our modified cross entropy loss function, while $\beta$ is 0.05 for the joint loss function $L_g$. 

The Adam solver \cite{Kingma2015a} is applied for both discriminator and generator with batch size of 8. The other three Adam parameters are set as default values, i.e. $\beta_1$ is 0.9, $\beta_2$ is 0.999 and $\epsilon$ is 1e-8. Learning rate is set to 0.0005 in the first epoch, decaying to 0.0001 in the following epochs. As we do not use dropout or batch normalization, the testing phase is exactly the same as training stage without reconfiguring network parameters. The whole network is trained on a single Titan X GPU from scratch. 

\subsection{Data Synthesis} %\label{sec:data}
For the task of 3D dense reconstruction from a single view, obtaining a large amount of training data is an obstacle. Existing real RGB-D datasets for surface reconstruction suffer from occlusions and missing data and there is no corresponding complete 3D structure for each single view. The recent work 3D-EPN \cite{Dai2017b} synthesizes data for 3D object completion, but their map encoding scheme is the complicated TSDF which is different from our network requirement. %Our networks only consume the output the simple occupancy voxel grids.

To tackle this issue, we use the ModelNet40 \cite{Wu2015} database to generate a large amount of training and testing data with synthetically rendered depth images and the corresponding complete 3D shape ground truth. Particularly, a subset of object categories is selected for our experiments. For each category, we generate training data from around 200 CAD models in the train folder, while synthesizing testing data from around 20 CAD models in the test folder. For each CAD model, we create a virtual depth camera to scan it from 125 different angles, 5 uniformly sampled views for each of roll, pitch and yaw space. For each virtual scan, both a depth image and the corresponding complete 3D voxelized structure are generated with regard to the same camera angle. That depth image is simultaneously transformed to a partial 2.5D voxel grid using virtual camera parameters. Then a pair of partial 2.5D view and the complete 3D shape is synthesized. Overall, around 20K training pairs and 2K testing pairs are generated for each 3D object category. All data are produced in Blender.

\section{Evaluation}\label{sec:eval}
In this section, we evaluate our 3D-RecGAN with comparison to alternative approaches and an ablation study to fully investigate the proposed network.

\subsection{Metrics}
We use two metrics to evaluate the performance of 3D reconstruction. The first metric is voxel Intersection-over-Union (IoU) between a predicted 3D voxel grid and its ground true voxel grid. It is formally defined as follows:
\begin{equation*}
IoU = \frac{\sum_{ijk} \left[  I (y_{ijk}^{'}>p) * I(y_{ijk}) \right] }{ \sum_{ijk} \left[  I \left( I(y^{'}_{ijk} >p) + I(y_{ijk}) \right) \right] } 
\end{equation*}
where $I()$ is an indicator function, (i,j,k) is the index of a voxel in three dimensions, $y^{'}_{ijk}$ is the predicted value at the (i,j,k) voxel, $y_{ijk}$ is the ground true value at (i,j,k), and $p$ is the threshold for voxelization. In all our experiments, p is set as 0.5. If the predicted value is over 0.5, it is more likely to be occupied from the probabilistic aspect. The higher the IoU value, the better the reconstruction of a 3D model. 

The second metric is the mean value of standard cross-entropy loss (CE) between a reconstructed shape and the ground true 3D model. It is formally presented as: 
\begin{equation*}
CE =\frac{1}{IJK} \sum_{ijk} \left[y_{ijk}\log(y^{'}_{ijk}) + (1 - y_{ijk})\log(1-y^{'}_{ijk})\right]
\end{equation*}
where $y^{'}_{ijk}$ and $y_{ijk}$ are the same as defined in above IoU, (I, J, K) are the voxel dimension sizes of output 3D models. The lower CE value is, the better 3D prediction.

\begin{figure*}[t]
	\setlength{\abovecaptionskip}{ 0 cm}
    \setlength{\belowcaptionskip}{ -8pt}
    \centering
    \includegraphics[width=1\textwidth]{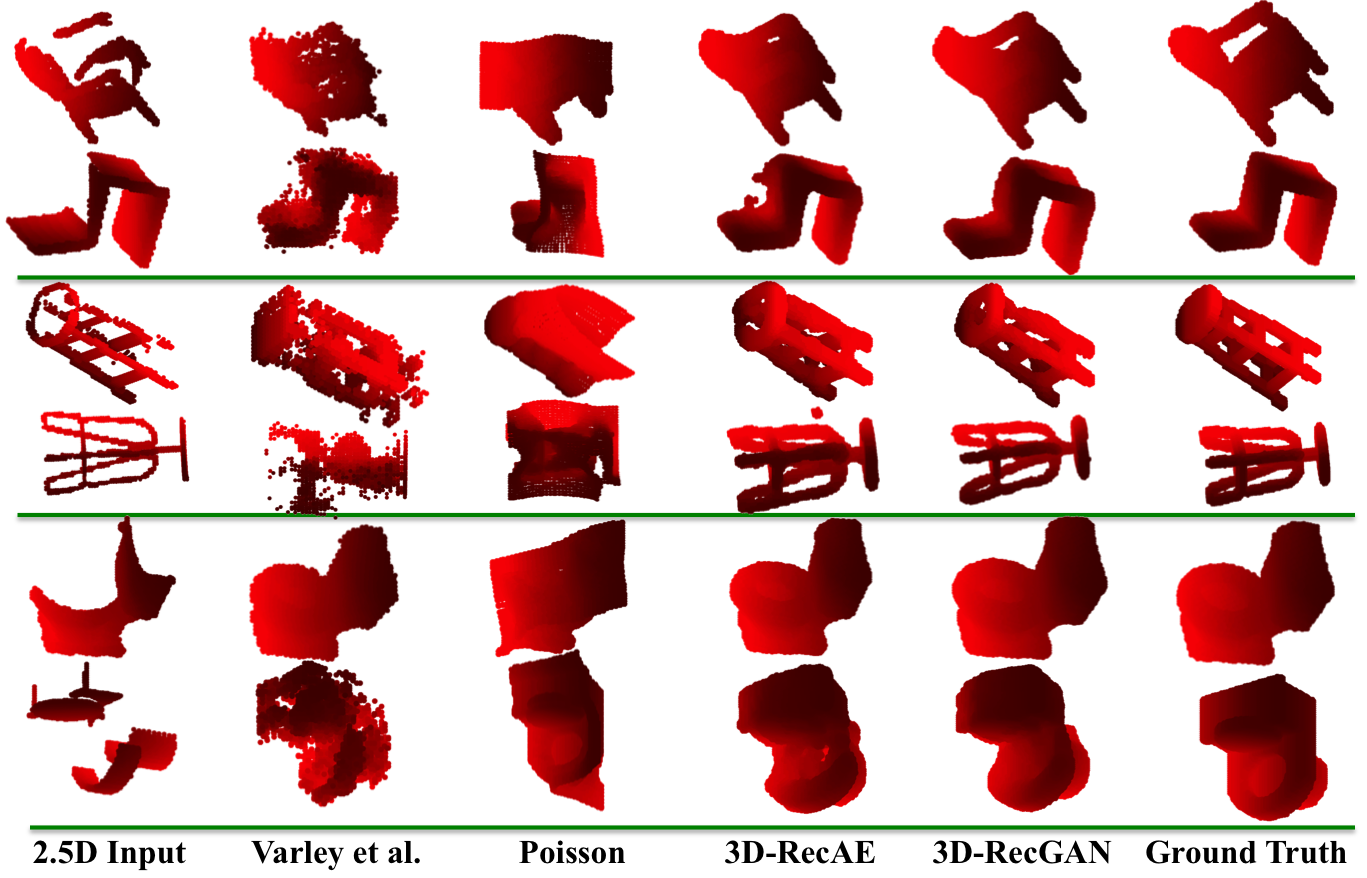}
    \caption{Qualitative results of per-category reconstruction from different approaches.}
    \label{fig:per_cat}
\end{figure*}
The above two metrics can evaluate the overall reconstruction performance, but the reconstructed geometric details are unlikely to be well evaluated in such way. Therefore, a large number of qualitative results from reconstructed 3D models are visualized in Section \ref{sec:comp}.

\subsection{Comparison}
\begin{figure*}[t]
	\setlength{\abovecaptionskip}{ 0 cm}
    \setlength{\belowcaptionskip}{ -8pt}
    \centering
    \includegraphics[width=1\textwidth]{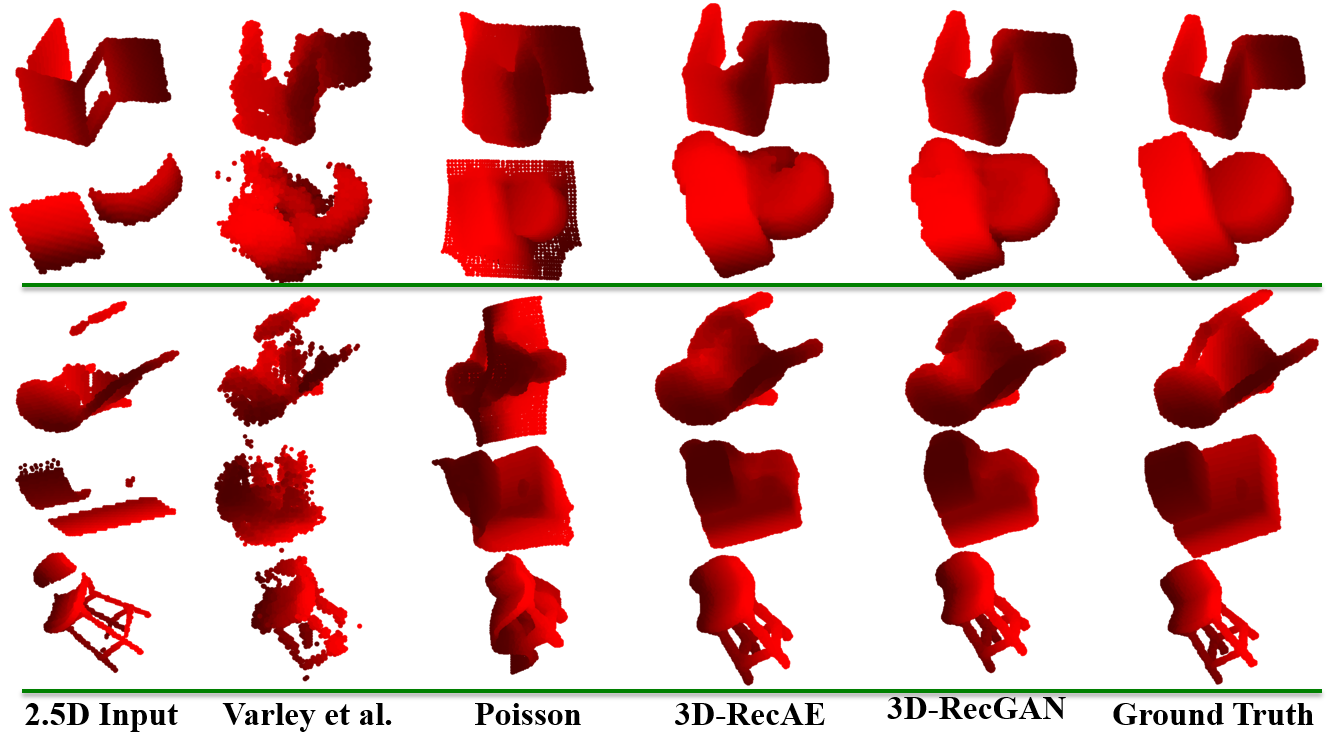}
    \caption{Qualitative results of multi-category reconstruction from different approaches.}
    \label{fig:multi_cat}
\end{figure*}
%%%%%%%%%%%%%%%%%%%%%%%%%%%%%%%%
\begin{figure*}[t]
	\setlength{\abovecaptionskip}{0 cm}
	\setlength{\belowcaptionskip}{-2.5pt}
	\begin{minipage}[t]{0.32\textwidth}
		\centerline{
			\includegraphics[width=1\textwidth]{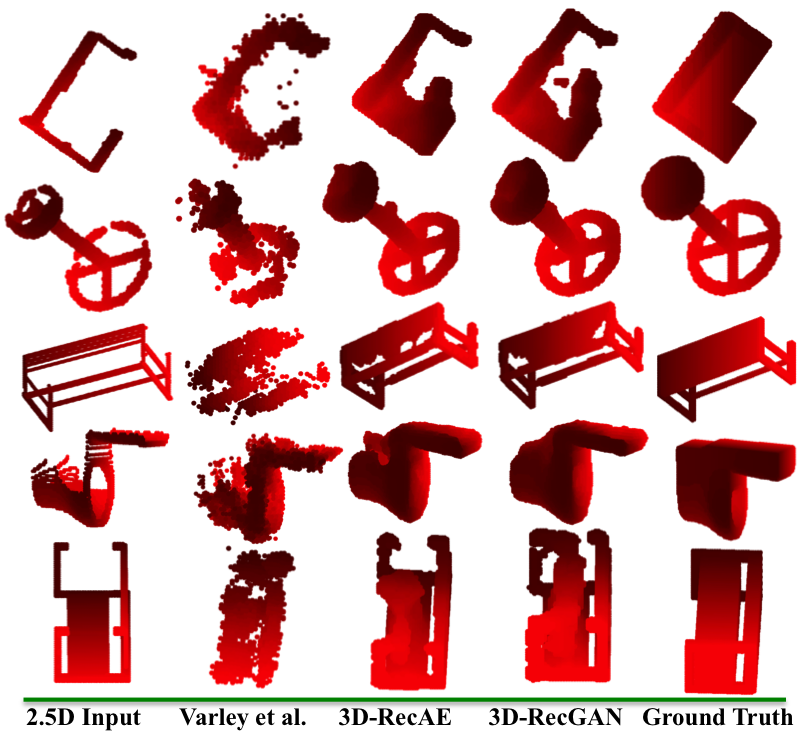}}
		\caption{Cross-category reconstruction results of Group 1.}
		\label{fig:cross_cat_chair}
	\end{minipage}
	\makebox[0.05in][]{}
	\begin{minipage}[t]{0.32\textwidth}
		\centerline{
			\includegraphics[width=1.0\textwidth]{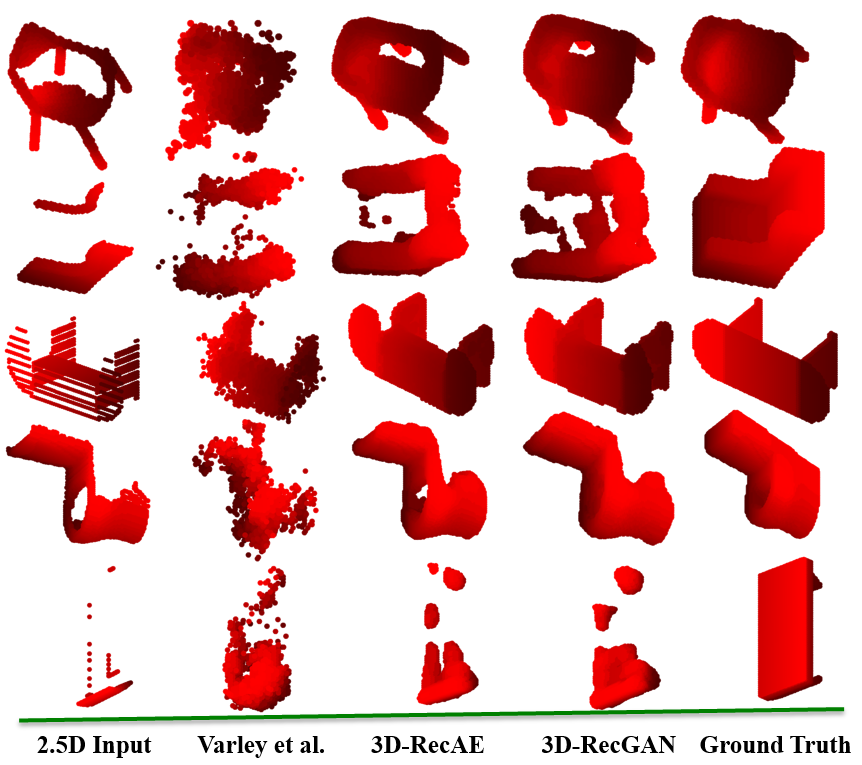}}
		\caption{Cross-category reconstruction results of Group 2.
		}\label{fig:cross_cat_stool}
	\end{minipage}
	\makebox[0.05in][]{}
	\begin{minipage}[t]{0.35\textwidth}
		\centerline{
			\includegraphics[width=1.0\textwidth]{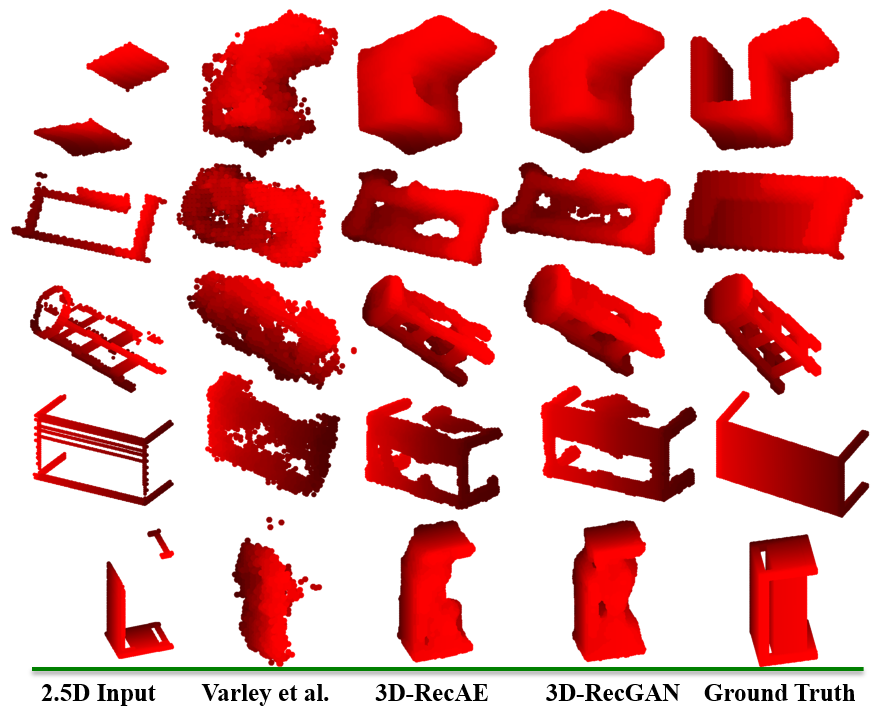}}
		\caption{Cross-category reconstruction results of Group 3.
		}\label{fig:cross_cat_toilet}
	\end{minipage}
\end{figure*}

We compare against two alternative reconstruction methods. The first is the well-known traditional Poisson surface reconstruction \cite{Kazhdan2006} \cite{Kazhdan2013}, which is mostly used for completing surfaces on dense point clouds. The second is the state-of-the-art deep learning based approach proposed by Varley et al. in \cite{Varley2017}, which is most similar to our approach in terms of input and output data encoding and the 3D completion task. It has encouraging reconstruction performance because of its two fully connected layers \cite{Long2015} in the model, but it is unable to deal with higher resolutions and it has less generality for shape completion. We also compare against the autoencoder alone in our network, i.e. without the GAN, named as 3D-RecAE for short.

(1) \textbf{Per-category Results}. The networks are separately trained and tested on three different categories with the same network configurations. Table \ref{tab:per_cat_iou_ce} shows the IoU and CE results, and Figure \ref{fig:per_cat} compares qualitative results from different reconstruction approaches.
\vspace{-0.2 cm}
\begin{table}[h]
\small
\caption{Per-category IoU and CE Loss.}
\centering
\label{tab:per_cat_iou_ce}
\tabcolsep=0.11cm
\begin{tabular}{|c|c|c|c|c|c|c|c|}
	\hline
       & \multicolumn{3}{c|}{IoU} & \multicolumn{3}{c|}{CE Loss} \\ \hline
  	 trained/tested on  & chair & stool & toilet & chair & stool & toilet \\ 
	\hlineB{3}
     Poisson & 0.180 & 0.189 & 0.150  & - & - & - \\
    \hline
    Varley \cite{Varley2017} &  0.564 & 0.273 & 0.503 &  0.132 & 0.189 & 0.177 \\
    \hline
    3D-RecAE &  0.633 & 0.488 & 0.520 & \textbf{0.069} & 0.085 & 0.166 \\
    \hline
    3D-RecGAN &  \textbf{0.661} & \textbf{0.501} & \textbf{0.569} &  0.074 & \textbf{0.083} & \textbf{0.157}  \\
    \hline
\end{tabular}
\end{table}

\vspace{-0.4 cm}
\begin{table}[h]
\small
\caption{Multi-category IoU and CE Loss.}
\centering
\label{tab:multi_cat_iou_ce}
\tabcolsep=0.11cm
\begin{tabular}{|c|c|c|c|c|}\hline
	 & \multicolumn{2}{c|}{IoU} & \multicolumn{2}{c|}{CE Loss} \\ \hline
    \begin{tabular}{@{}c@{}}trained/ \\ tested on\end{tabular}  & chair/toilet & \begin{tabular}{@{}c@{}}chair/toilet \\ /stool\end{tabular} 
     
     & chair/toilet & \begin{tabular}{@{}c@{}}chair/toilet \\ /stool\end{tabular} \\
	\hlineB{3}
    Poisson & 0.165 & 0.173 & - & -  \\ \hline
    Varley \cite{Varley2017} &  0.493 & 0.453 &  0.125 & 0.173    \\
    \hline
    3D-RecAE &  0.514 & 0.487 &  0.127 & 0.109 \\
    \hline
    3D-RecGAN &  \textbf{0.554} & \textbf{0.513} &  \textbf{0.117} & \textbf{0.101} \\
    \hline
\end{tabular}
\end{table}

\vspace{-0.25 cm}
(2) \textbf{Multi-category Results}. To study the generality, the networks are trained and tested on multiple categories without given any class labels. Table \ref{tab:multi_cat_iou_ce} shows the IoU and CE results, and Figure \ref{fig:multi_cat} shows the qualitative results.

(3) \textbf{Cross-category Results}. To further investigate the generality, the network is trained on one category, but tested on another five different categories. Particularly, in Group 1, the network is trained on chair, tested on sofa, stool, table, toilet, and TV stand; in Group 2, the network is trained on stool, tested on chair, sofa, table, toilet, and TV stand; in Group 3, the network is trained on toilet, tested on chair, sofa, stool, table, and TV stand. Table \ref{tab:cross_cat_iou_ce} shows the IoU and 
CE results; Figure \ref{fig:cross_cat_chair}, \ref{fig:cross_cat_stool} and \ref{fig:cross_cat_toilet} compare qualitative cross-category reconstruction results of Group 1, Group 2 and Group 3 respectively. 
\vspace{-0.25 cm}
\begin{table}[h]
\small
\caption{Cross-category IoU and CE Loss.}
\centering
\label{tab:cross_cat_iou_ce}
%\begin{tabularx}{\linewidth}{|L|c|L|} 
\tabcolsep=0.06cm
\begin{tabular}{|c|c|c|c|c|c|c|}\hline
	& \multicolumn{3}{c|}{IoU} & \multicolumn{3}{c|}{CE Loss} \\ \hline
     & Group1 & Group2 & Group3  & Group1 & Group2 & Group3 \\ \hline
    \hlineB{3}
    Varley \cite{Varley2017} &  0.253 & 0.221 & 0.277 &  0.430 & 0.425 & 0.297  \\
    \hline
    3D-RecAE &  0.353  & 0.362 & 0.349 &  \textbf{0.218} & \textbf{0.117} & \textbf{0.149} \\
    \hline
    3D-RecGAN & \textbf{0.356} & \textbf{0.369} & \textbf{0.351} &  0.264 & 0.345 & 0.162  \\
    \hline
\end{tabular}
\end{table}

\vspace{-0.25 cm}
Overall, the above extensive experiments for per-category and multi-category object reconstruction demonstrate that our proposed 3D-RecGAN is able to complete partial 2.5D views with accurate structures and fine-grained details, outperforming the state of the art by a large margin. In addition, our 3D-RecGAN performs well in the challenging cross-category reconstruction task, which demonstrates that our novel model implicitly learns the geometric features and their correlations among different object categories. 

\eat{
\begin{figure}[h]
	\setlength{\abovecaptionskip}{ 0 cm}
    \setlength{\belowcaptionskip}{ -8pt}
    \centering
    \includegraphics[width=0.5\textwidth]{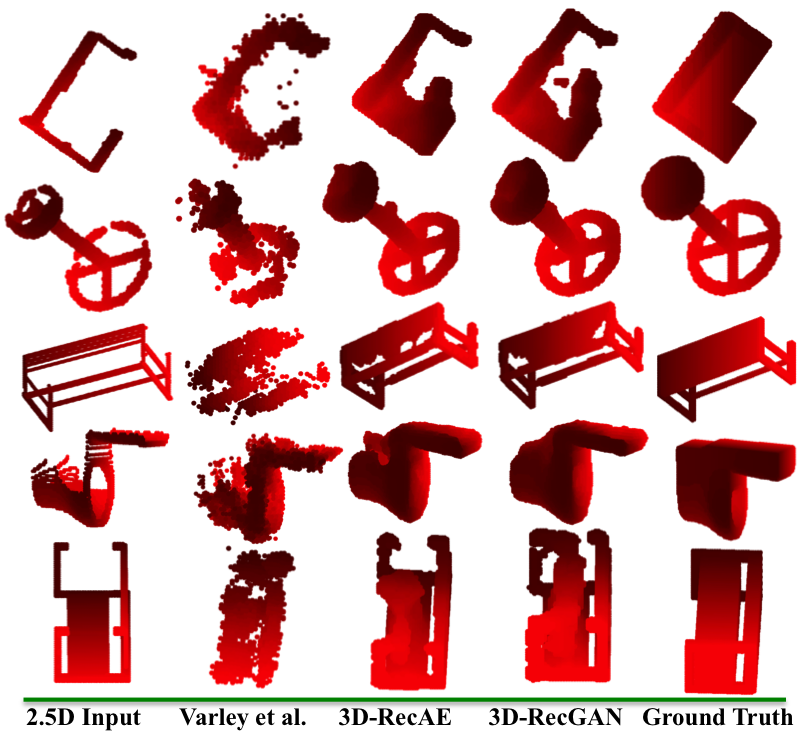}
    \caption{Reconstruction results of Group 1.}
    \label{fig:cross_cat_chair}
\end{figure}
\begin{figure}[h]
	\setlength{\abovecaptionskip}{ 0 cm}
    \setlength{\belowcaptionskip}{ -8pt}
    \centering
    \includegraphics[width=0.5\textwidth]{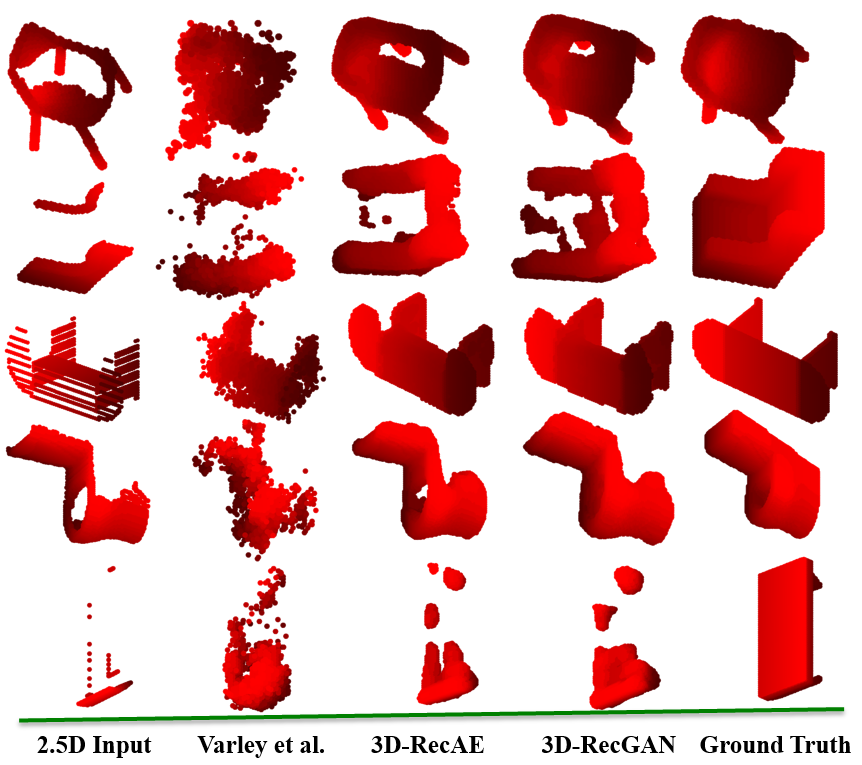}
    \caption{Reconstruction results of Group 2.}
    \label{fig:cross_cat_stool}
\end{figure}
\begin{figure}[h]
	\setlength{\abovecaptionskip}{ 0 cm}
    \setlength{\belowcaptionskip}{ -8pt}
    \centering
    \includegraphics[width=0.5\textwidth]{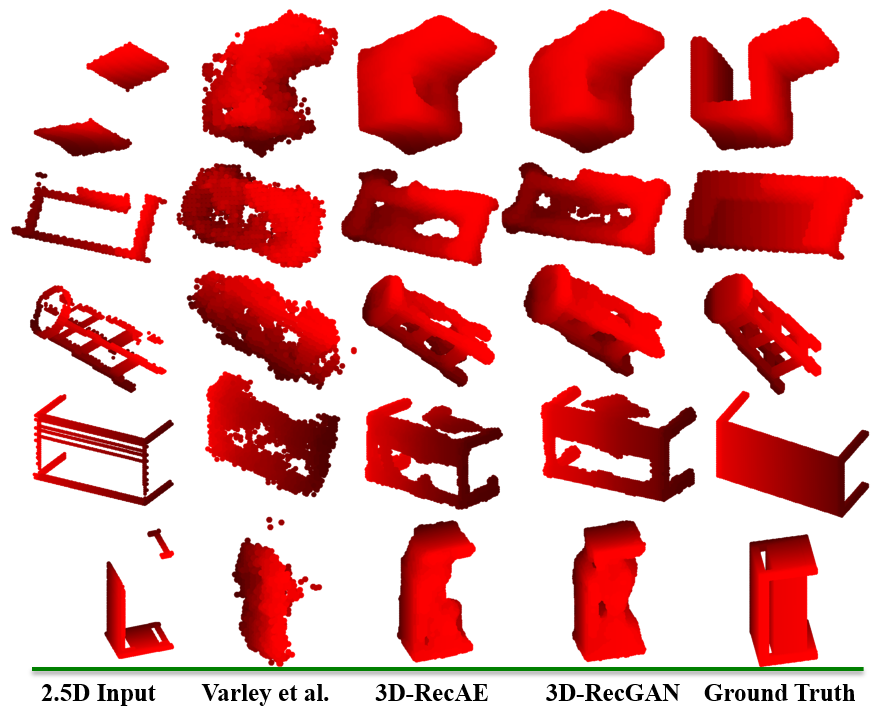}
    \caption{Reconstruction results of Group 3.}
    \label{fig:cross_cat_toilet}
\end{figure}
}\label{sec:comp}

\section{Conclusion}\label{sec:sum}
In this work, we proposed a novel framework 3D-RecGAN that reconstructs the full 3D structure of an object from an arbitrary depth view. By leveraging the generalization capabilities of autoencoders and generative networks, our 3D-RecGAN predicts accurate 3D structures with fine details, outperforming the traditional Poisson algorithm and the method in Varley et al.\cite{Varley2017} in single-view shape completion for individual object category. We further tested our network's ability to perform reconstruction on multiple categories without providing any object class labels during training or testing, and it showed that our network is able to predict satisfactory 3D shapes. Finally, we investigated the network's reconstruction performance on unseen categories of objects. We showed that even in very challenging cases, the proposed approach can still predict plausible 3D shapes. This confirms that our network has the capability of learning general 3D latent features of the objects, rather than simply fitting a function for the training datasets. In summary, our network only requires a single depth view to recover an accurate complete 3D shape with fine details.

{\small
\bibliographystyle{ieee}
\bibliography{Mendeley}
}
\end{document}